\pdfoutput=1

\documentclass[11pt]{article}

\usepackage[preprint]{acl}

\usepackage{times}
\usepackage{latexsym}

\usepackage[T1]{fontenc}

\usepackage[utf8]{inputenc}

\usepackage{microtype}

\usepackage{inconsolata}

\usepackage{graphicx}
\usepackage{amsmath,amsfonts,amssymb}
\usepackage{array}
\usepackage[caption=false,font=normalsize,labelfont=sf,textfont=sf]{subfig}
\usepackage{textcomp}
\usepackage{stfloats}
\usepackage{url}
\usepackage{verbatim}
\usepackage{graphicx}
\usepackage[affil-it]{authblk}
\usepackage{balance}
\usepackage{caption}

\usepackage{booktabs}
\usepackage{mathabx}
\usepackage{algorithm}
\usepackage[noend]{algpseudocode}
\usepackage{multirow}
\usepackage{wrapfig}

\usepackage{hyperref}
\usepackage{enumitem}

\usepackage{amsmath,amssymb,amsfonts}
\usepackage{algorithm}
\usepackage{algpseudocode}
\usepackage{graphicx}
\usepackage{textcomp}
\usepackage{xcolor}
\usepackage{booktabs}
\usepackage{siunitx}
\usepackage{multirow}
\usepackage{array}
\usepackage{amssymb}
\usepackage{makecell}
\usepackage{comment}
\usepackage{threeparttable}
\newtheorem{theorem}{Theorem}
\usepackage[utf8]{inputenc}
\usepackage{pifont}
\DeclareUnicodeCharacter{2713}{\checkmark}  
\DeclareUnicodeCharacter{2717}{\ding{55}}   
\definecolor{acceptgreen}{RGB}{0,180,0}  
\definecolor{rejectred}{RGB}{255,0,0}
\definecolor{oursblue}{RGB}{40,100,180}

\newcommand{\best}[1]{\textbf{#1}}

\title{PipeSpec: Breaking Stage Dependencies in Hierarchical LLM Decoding}

\author{
  \textbf{Bradley McDanel}\textsuperscript{1} \quad
  \textbf{Sai Qian Zhang}\textsuperscript{2} \quad
  \textbf{Yunhai Hu}\textsuperscript{2} \quad
  \textbf{Zining Liu}\textsuperscript{3} \quad
\\
  \textsuperscript{1}Franklin and Marshall College \quad
  \textsuperscript{2}New York University \quad
  \textsuperscript{3}University of Pennsylvania
\\

{{\texttt{{bmcdanel@fandm.edu} \quad \{yunhai.hu, sai.zhang\}@nyu.edu}}} \\
  {\texttt{{zliu0@seas.upenn.edu}}}
}

\begin{document}

\maketitle

\begin{abstract}
Speculative decoding accelerates large language model inference by using smaller draft models to generate candidate tokens for parallel verification. However, current approaches are limited by sequential stage dependencies that prevent full hardware utilization. We present PipeSpec, a framework that generalizes speculative decoding to $k$ models arranged in a hierarchical pipeline, enabling asynchronous execution with lightweight coordination for prediction verification and rollback. Our analytical model characterizes token generation rates across pipeline stages and proves guaranteed throughput improvements over traditional decoding for any non-zero acceptance rate. We further derive closed-form expressions for steady-state verification probabilities that explain the empirical benefits of pipeline depth. Experimental results show that PipeSpec achieves up to 2.54$\times$ speedup while outperforming state-of-the-art methods. We validate PipeSpec across text summarization and code generation tasks using LLaMA 2 and 3 models, demonstrating that pipeline efficiency increases with model depth, providing a scalable approach to accelerating LLM inference on multi-device systems.
\end{abstract}

\section{Introduction}
Large language models (LLMs) have transformed natural language processing through their remarkable ability to understand and generate human-like text. However, the fundamental requirement of autoregressive token generation, where each token must be generated sequentially based on all previous tokens, creates significant performance bottlenecks. This limitation is particularly pronounced in modern LLMs with 100B or more parameters~\cite{dubey2024llama}, making real-time applications challenging. Recent advances in speculative decoding have shown promise by leveraging smaller, faster models to draft candidate tokens for verification by larger models. However, current approaches still face fundamental efficiency limits due to their strict sequential dependencies between draft and verification stages.

\begin{figure}
\centering
\includegraphics[width=\linewidth]{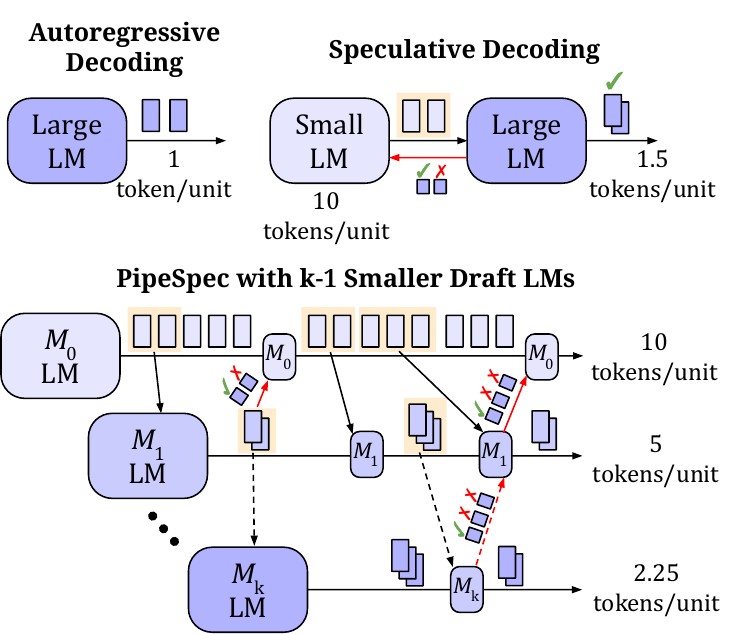}
\caption{Comparison of different LLM decoding approaches. Top Left: Traditional autoregressive decoding (1 token/unit). Top Right: Speculative decoding using a small draft model (10 tokens/unit) for parallel verification by a large model (1.5 tokens/unit). Bottom: Our PipeSpec framework with $k-1$ draft models in a pipeline feeding into the large model ($M_k$), achieving 2.25 tokens/unit through pipelined parallelism. Checkmarks (\textcolor{acceptgreen}{✓}) show accepted predictions while crosses (\textcolor{rejectred}{✗}) indicate rejections triggering pipeline rollbacks.}
\label{fig:overview}
\end{figure}

As illustrated in Figure~\ref{fig:overview}, traditional autoregressive decoding using a single large model is limited to 1 token per unit time due to strict sequential dependencies. Standard speculative decoding improves throughput to 1.5 tokens per unit time by employing a small draft model (10 tokens/unit) to generate candidates for batch verification by the large model. However, this approach still suffers from alternating idle periods where either the draft or verify model must wait for the other to complete.

Our key insight is that these limitations can be overcome through pipelining of multiple models. PipeSpec introduces a novel $k$-model architecture where each consecutive pair of models operates in an asynchronous producer-consumer relationship. In the three-model configuration shown in Figure~\ref{fig:overview} (bottom), an initial small model ($M_0$) rapidly generates draft tokens (10 tokens/unit), which are progressively refined by a medium-sized model ($M_1$, 5 tokens/unit) before final verification by the large model ($M_k$, 2.25 tokens/unit). This hierarchical structure provides two key advantages: (1) each stage operates asynchronously, enabling continuous parallel execution without idle periods, and (2) the intermediate models provide higher-quality draft tokens compared to single-draft approaches while still benefiting from their own draft-verify speedups.

PipeSpec operates through optimistic execution, where each model generates tokens assuming downstream acceptance. When a model rejects a prediction (marked as \textcolor{rejectred}{✗}), it triggers a rollback cascade -- all subsequent predictions in earlier pipeline stages must be discarded and regenerated. This enables PipeSpec to maintain higher throughput than traditional Speculative Decoding. The main contributions of this work are:
\begin{itemize}[noitemsep, topsep=0pt]
    \item A novel hierarchical pipeline architecture for speculative decoding that breaks traditional stage dependencies, enabling continuous parallel execution across $k$ models of increasing size and accuracy
    \item An analytical model that derives expected token generation rates and steady-state verification probabilities for pipelined models, with a proof of improved throughput over autoregressive decoding
    \item A complete multi-GPU implementation with efficient inter-device communication and rollback mechanisms, validated through extensive experiments showing consistent speedup over existing state-of-the-art speculative decoding approaches
\end{itemize}

\section{Related Work}
\label{sec:bg}

\subsection{LLM Inference Acceleration}
LLM inference consists of two distinct computational phases: prefill and decode. The prefill phase processes the initial input prompt, computing attention across all input tokens with quadratic memory scaling. The decode phase generates new tokens sequentially, requiring attention computation only against previous tokens' cached key-value pairs, making it more computationally bounded than memory bounded.

Recent research has targeted hardware-level optimizations for both phases. For prefill, FlashAttention~\cite{dao2022flashattention, dao2023flashattention} optimizes attention computation through tiling and recomputation strategies, particularly important for long sequences where naive implementations would exceed GPU memory bandwidth. Other approaches focus on GPU utilization~\cite{hong2023flashdecoding++, Vaidya2023TensorRTLLM, patel2024splitwise} and efficient key-value cache management~\cite{aminabadi2022deepspeed, sheng2023flexgen, kwon2023efficient}. While these approaches optimize individual model execution, they are complementary to our proposed PipeSpec framework, which focuses on algorithmic speedups through pipelined speculative execution.

\subsection{Speculative Decoding}

While prefill optimizations like FlashAttention address the initial prompt processing, speculative decoding targets the decode phase bottleneck by leveraging parallel verification. First proposed by Stern et al.~\cite{stern2018blockwise}, the core idea is to use a smaller, faster draft model to generate multiple tokens sequentially that are then verified in parallel by the larger model, amortizing the cost of loading model weights and KV cache across multiple tokens (see Figure~\ref{fig:overview} top right).

Building on this foundation, researchers have developed various approaches to improve the efficiency of this draft-verify process. Tree-structured verification approaches~\cite{miao2024specinfer,li2024eagle,fu2024break} expand beyond single-path prediction to explore multiple candidate sequences simultaneously, increasing the likelihood of successful verification of draft tokens. Other techniques like token distillation~\cite{zhoudistillspec}, layer skipping~\cite{zhang2023draft,elhoushi2024layer}, and retrieval-augmented drafting~\cite{he2024rest} aim to enhance draft model quality while maintaining low computational overhead. The MEDUSA framework~\cite{cai2024medusa} introduced specialized decoding heads to improve drafting efficiency without requiring a separate draft model; notably, all these algorithmic approaches are orthogonal to and could be combined with our systems-level pipeline optimization strategy.

More recently, several approaches have explored using multiple draft models to further accelerate inference. TRIFORCE~\cite{sun2024triforce} focuses specifically on extremely long-sequence generation (e.g., 100k context windows) by using the original model with partial KV cache as an intermediate draft stage. Spector and R\'{e}~\cite{spector2023accelerating} explored tree-structured batches across multiple draft models, though their approach remains tied to synchronous execution between stages. Our evaluation in Section~\ref{sec:eval} includes tiered speculative decoding configurations (using multiple draft models in sequence) which capture some of these benefits, but PipeSpec's key innovation is introducing true asynchronous pipelining where each model pair operates independently in a producer-consumer relationship. This fundamental architectural difference enables significantly higher throughput by maximizing hardware utilization across all available models, as demonstrated in our results.

\section{Pipelined Speculative Decoding}
\label{sec:pipe-spec}

In this section, we first describe the operation of Pipelined Speculative Decoding (PipeSpec) as a $k$-stage pipeline (Section \ref{sec:pipe-spec:overview}). We then present the core algorithm of PipeSpec (Section \ref{sec:pipe-spec:algorithm}). Finally, we develop a theoretical framework to analyze PipeSpec's performance characteristics and compare it with existing approaches (Section \ref{sec:pipe-spec:optimization}).

\subsection{Overview}
\label{sec:pipe-spec:overview}

\begin{figure*}[!htb]
    \centering
    \includegraphics[width=1\linewidth]{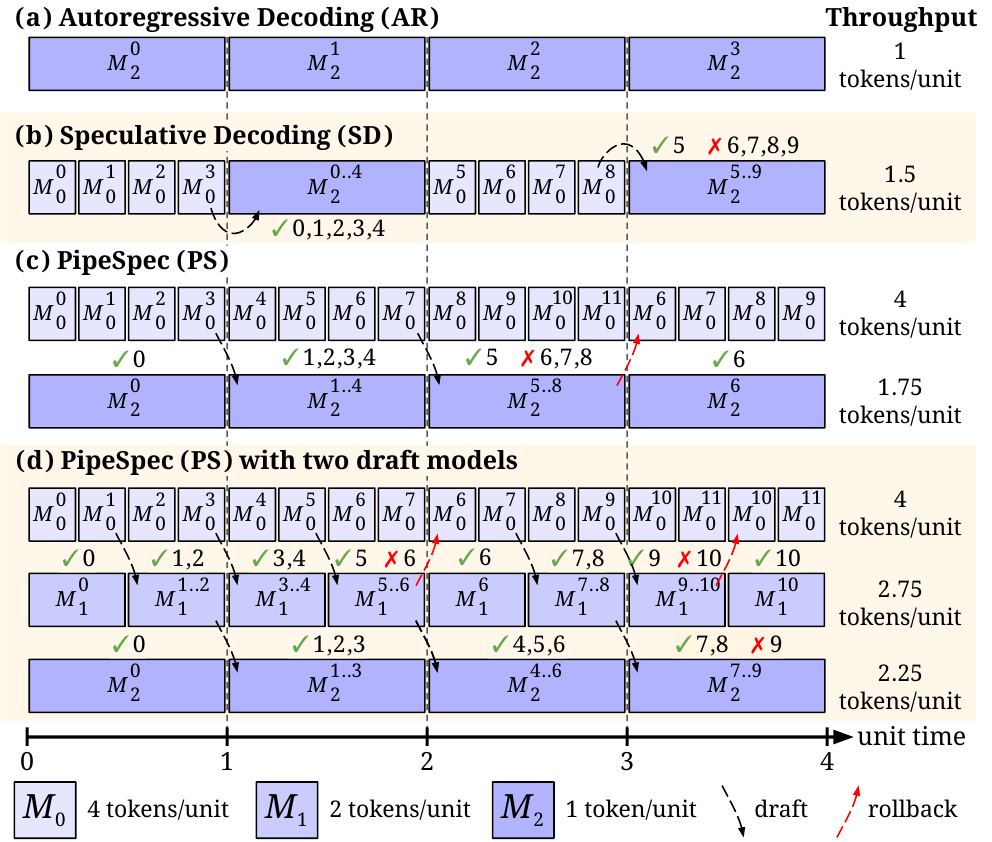}
    \caption{Comparison of different decoding approaches showing token generation over time. From top to bottom: (1) Traditional autoregressive decoding (AR) with sequential token generation using a single model $M_2$, (2) Standard speculative decoding (SD) using a draft model $M_0$ to generate candidate tokens verified in batches by $M_2$, (3) PipeSpec (PS) with 2 models showing continuous parallel execution between $M_0$ and $M_2$, and (4) PipeSpec with 3 models demonstrating hierarchical speculation across \{$M_0$, $M_1$, $M_2$\}.}
    \label{fig:decoding-comp}
\end{figure*}

Figure~\ref{fig:decoding-comp} compares token generation across different decoding approaches. The simplest approach, autoregressive decoding (top), uses a single large model ($M_2$) to generate tokens one at a time, achieving a throughput of 1 token per time unit. Traditional speculative decoding (second row) improves upon this by using a small draft model ($M_0$) that can generate tokens 4 times faster than $M_2$. However, despite this theoretical speedup, two key limitations prevent the system from achieving its full potential:

\begin{enumerate}[noitemsep, topsep=0pt, parsep=0pt, partopsep=0pt]
   \item \textbf{Synchronous Execution:} The draft and verify stages operate in strict lockstep---$M_0$ must wait for $M_2$ to complete verification before generating the next batch of tokens. This creates alternating idle periods where $M_0$ is blocked waiting for verification results, and periods where $M_2$ is idle while new draft tokens are generated.
   
   \item \textbf{Misprediction Penalty:} When $M_2$ rejects a prediction (marked with \textcolor{rejectred}{✗} in the figure), all subsequent draft tokens in that batch become invalid and must be discarded. For example, in Figure~\ref{fig:decoding-comp}(b), the rejection of token 6 invalidates the draft work done for tokens 7, 8, and 9, incurring a significant misprediction penalty.
\end{enumerate}

These limitations combine to reduce the effective throughput to 1.5 tokens per unit, far below the theoretical maximum of the draft model.

PipeSpec introduces two key architectural innovations to address these inefficiencies. First, in its basic two-model configuration, we eliminate artificial synchronization requirements between draft and verification stages. This allows $M_0$ to optimistically generate additional draft tokens while $M_2$ verifies the prior batch of tokens in parallel. Assuming all tokens are accepted, the next verification stage can start immediately with a new batch of tokens, leading to improved throughput. However, when draft predictions are rejected, the system still needs to trigger a targeted rollback (shown by red dashed lines) and resumes generation from the last valid token.

While this two-model configuration addresses the synchronization problem, misprediction penalties still impact performance significantly. To mitigate this, we introduce a three-model configuration with an intermediate model ($M_1$) that reduces misprediction penalties in two ways: (1) it quickly filters out low-quality predictions from $M_0$ before they reach the expensive $M_2$ verification stage, and (2) it provides $M_2$ with higher-quality draft tokens that are less likely to be rejected. This hierarchical refinement enables $M_1$ to serve as both a lightweight verification stage for $M_0$ and an improved draft model for $M_2$, achieving 2.25 tokens per unit (9 tokens verified in 4 time units) in Figure~\ref{fig:decoding-comp}(d) while maintaining continuous parallel execution. This pipeline structure naturally extends to additional stages, with each intermediate model further reducing misprediction penalties through progressive refinement.

\subsection{Algorithm}
\label{sec:pipe-spec:algorithm}

Algorithm \ref{alg:pipespec} presents the core mechanism of Pipelined Speculative Decoding (PipeSpec). Each model $i$ in our $K$-model pipeline maintains its own output buffer $O_i$, operating asynchronously while coordinating through a lightweight rejection mechanism. The first model ($i=0$) continuously generates draft tokens, while verification models ($i>0$) compare incoming draft tokens against their own token predictions. When a verification model rejects tokens (due to prediction mismatch), it signals earlier stages to rollback their buffers $O_i$ to maintain consistency. The pipeline terminates when the final model $O_K$ generates an end token, ensuring all tokens have been properly verified through the complete pipeline.

\begin{algorithm}[!htb]
\footnotesize 
\caption{Pipelined Speculative Decoding}
\label{alg:pipespec}
\begin{algorithmic}[1]
\Require Input prompt, Models $[M_0...M_K]$
\Ensure Generated sequence $O_K$
\State Let $O_i$ be token buffer for model $i$ with length $|O_i|$
\While{not finished generating}
    \For{each model $i$ running in parallel}
        \If{received rejection from stage $j > i$}
            \State Rollback $O_i$ to match $O_j$'s last token
        \EndIf
        \If{$i = 0$}  \Comment{First generates drafts}
            \State Generate next token, append to $O_0$
        \Else \Comment{Others verify drafts}
            \State Get draft tokens from $O_{i-1}$ 
            \State Generate token predictions
            \State Compare against predicted tokens
            \State Append matching tokens to $O_i$
            \If{any tokens do not match predictions}
                \State Signal rejection to earlier stages
            \EndIf
        \EndIf
    \EndFor
    \If{end token in $O_K$}
        \State \textbf{break}
    \EndIf
\EndWhile
\State \Return $O_K$
\end{algorithmic}
\end{algorithm}

\subsection{Theoretical Performance Analysis}
\label{sec:pipe-spec:optimization}

Let \(\mathcal{M} = \{M_0, M_1, \dots, M_K\}\) represent a collection of LLMs ordered in increasing size, with \(t_i\) denoting the per-token generation time for model \(M_i\). For any consecutive pair of models \(M_i\) and \(M_{i+1}\), the token acceptance rate \(\alpha_{i,i+1}\) is the probability that tokens generated by \(M_i\) are accepted by \(M_{i+1}\) during verification. In a hierarchical speculative decoding framework with \(K\) stages, the draft model \(\mathcal{M}_{draft}\) can be any model from \(\{M_0, \dots, M_{K-1}\}\), while the target model \(\mathcal{M}_{target}\) is \(M_K\). The expected number of tokens $N(M_{i})$ generated at $M_{i}$ at each decoding step is then defined as:
\begin{equation}
    \label{eqn:iterative}
    E(N\left(M_{i}\right)) = (1 - \rho_{i}) \cdot 1 + \rho_{i} \cdot \frac{1 - \alpha_{i-1,i}^{\gamma_i+1}}{1 - \alpha_{i-1,i}}
\end{equation}
where \(\rho_i\) represents the probability that \(M_i\) verifies the draft tokens generated by \(M_{i-1}\) in the window. \(\gamma_i\) represents the token window size of \(M_i\), and \(\alpha_{i-1,i}\) is the probability that a token from \(M_{i-1}\) is successfully verified by \(M_i\). If any draft token is rejected, the verification model generates one token in the next step, as illustrated in the PipeSpec workflow in Figure~\ref{fig:decoding-comp}. Otherwise, $\frac{1 - \alpha_{i-1,i}^{\gamma_i+1}}{1 - \alpha_{i-1,i}}$ tokens will be produced, as derived from~\cite{leviathan2023fastinferencetransformersspeculative}.

\(\rho_i\) should represent the probability of a steady state, because its calculation needs to take into account all previous token generation conditions of \(M_i\) up to the current step. To enter the verification process, one of the following two conditions must be met: if no verification was performed last time, the first draft token to be verified generated by $M_{i-1}$ is consistent with the new token generated by $M_{i}$ last time. Alternatively, if verification was performed last time, all draft tokens must pass, then the first draft token to be verified is consistent with the new token generated by $M_{i}$ in the last step. Given \(T = (t_0, t_1, \ldots, t_n)\), where \(t_j\) represents the \(j\)-th token generation step, the model \(M_i\) has performed calculations up to time step \(t_j\). \(\rho_i(t_j)\) represents the probability that $M_i$ will do verification at its $j$-th time step, which satisfies the following recursive condition

\begin{equation}
\label{eqn:recursive}
\rho_{i}(t_j) = \rho_{i}(t_{j-1}) \cdot \alpha_{i-1,i}^{\gamma_i+1} + (1 - \rho_{i}(t_{j-1})) \cdot \alpha_{i-1,i}
\end{equation}
for $j > 1$, and we also have $\rho_{i}(t_0) = \alpha_{i-1,i}$.

According to the recursive equation~\ref{eqn:recursive}, when (\(n \to \infty\)), $\rho_i$ reaches its stable state, which is given by, 

\begin{equation}
\rho_{i} = \lim_{n \to \infty} \frac{1}{n+1} \sum_{j=0}^{n} \rho_i(t_j) = \frac{\alpha_{i-1,i}}{1 - \alpha_{i-1,i}^{\gamma_i + 1} + \alpha_{i-1,i}}
\end{equation}

\begin{theorem}
For any \(0 < \alpha < 1\) and \(0 < \gamma\), the PipeSpec scheme generates a higher number of tokens per step.
\end{theorem}

\begin{equation}
    PipeSpec(P) = (1 - \rho_{k}) \cdot 1 + \rho_{k} \cdot \frac{1 - \alpha_{k-1,k}^{\gamma_k + 1}}{1 - \alpha_{k-1,k}} > 1
\end{equation}

It is obvious that $PipeSpec(P)$ is greater than 1 for any $\alpha$ and $\gamma$ greater than 0, so the pipeline specification is definitely better than autoregressive decoding.

As for standard speculative decoding, we assume a two stage configuration $P_a = (M_{d}, M_{t})$, where $M_{d}$ represents draft model, \(M_t\) represents verification model. The acceptance rate for \( M_d \) and \( M_t \) is represented by \(\alpha_{d,t}\), while the window size is given as \(\gamma_t\). Additionally, \( c_{d,t} \) denotes the speed ratio between the two models. Since each verification requires waiting for \(M_d\) to generate \(\gamma_t\) draft tokens, an additional \(\frac{\gamma_t}{c_{d,t}}\) units of time are spent generating these draft tokens.
Therefore, we can obtain the theoretical speedup of standard speculative decoding.

\begin{equation}
    SD(P_a) = \frac{1 - \alpha_{d,t}^{\gamma_t + 1}}{(1 - \alpha_{d,t}) \left( \frac{\gamma_t}{c_{d,t}} + 1 \right)}
\end{equation}

If $\alpha_{d,t}$ is low, standard speculative decoding performs worse than autoregressive decoding due to the combined overhead of waiting for draft tokens and frequent verification failures. PipeSpec outperforms standard speculative decoding in this scenario since it eliminates waiting times through asynchronous execution. When $\alpha_{d,t}$ approaches its ideal case (higher acceptance rates), PipeSpec's theoretical performance improvement can be approximated as:

\begin{equation}
    PipeSpec(P_a) \approx \frac{1 - \alpha_{d,t}^{\gamma_t+1}}{1 - \alpha_{d,t}}
\end{equation}

Since PipeSpec does not need to spend time waiting for the draft models to generate draft tokens, it clearly has better performance than standard speculative decoding. This theoretical analysis aligns with our empirical observations in Section~\ref{sec:eval}, where we see PipeSpec achieving 2.54$\times$ speedup with a three-model configuration on LLaMA3.1-70B compared to 1.32$\times$ for traditional speculative decoding. The relationship between acceptance rate $\alpha_{d,t}$ and throughput is particularly evident in our HumanEval results, where the \{1B, 8B, 70B\} pipeline demonstrates how intermediate model refinement improves acceptance rates. For example, when using the 8B model as an intermediate verifier, we observe an additionaly 12\% speedup due to an increase in acceptance rates for tokens reaching the 70B model compared to direct 1B$\rightarrow$70B verification. This empirical improvement validates our theoretical prediction that pipeline depth correlates positively with efficiency gains, as each intermediate stage acts as both a verification filter and an improved draft model for subsequent stages.

\section{Evaluation}
\label{sec:eval}
Our evaluation examines four aspects: end-to-end performance across summarization and code generation tasks (\ref{sec:eval:perf}), token acceptance patterns and timing characteristics (\ref{sec:eval:token-analysis}), the impact of lookahead window sizes on throughput (\ref{sec:eval:lookahead}), and GPU resource utilization (\ref{sec:eval:util}).

\subsection{Experimental Setup}
\label{sec:eval:setup}
All experiments were conducted on four NVIDIA A100-40GB GPUs interconnected via NVLink. GPU performance metrics were collected using nvidia-smi with 100ms sampling intervals.

We evaluated on the CNN/DM~\cite{nallapati2016abstractive} and XSUM~\cite{narayan2018dont} text summarization datasets, and the HumanEval~\cite{chen2021evaluating} code generation benchmark. For models, we employed LLaMA-2~\cite{touvron2023llama} and LLaMA-3~\cite{dubey2024llama} variants, with each model allocated to dedicated GPU(s). The 70B variants used 4-bit quantization and were split across 2 GPUs. All experiments used greedy decoding (temperature=0.0) with maximum sequence length of 512 tokens, following prior work~\cite{zhang2023draft,elhoushi2024layer}, to ensure a fair comparison.

\subsection{Performance Analysis}
\label{sec:eval:perf}

Table~\ref{tab:results} demonstrates the performance advantages of PipeSpec across multiple datasets and model configurations. The notation \{M$_0$, M$_1$, ..., M$_k$\} in the Models column denotes a pipeline of models where M$_0$ is the smallest/fastest model and M$_k$ is the verifier model. In traditional speculative decoding, these models operate sequentially -- each model must wait for draft tokens from the previous model before beginning generation. In contrast, PipeSpec allows these models to operate asynchronously as discussed earlier in Section~\ref{sec:pipe-spec:overview}. Our evaluation reveals several significant trends:

First, PipeSpec consistently outperforms standard speculative decoding when using identical model configurations. For example, with a \{68M, 7B\} configuration on CNN/DM, PipeSpec achieves a 1.40$\times$ speedup compared to 1.35$\times$ for standard speculative decoding. This advantage becomes more pronounced with larger models - on HumanEval using LLaMA3.1-70B, PipeSpec with \{8B, 70B\} achieves 2.27$\times$ speedup versus 1.32$\times$ for speculative decoding.

Second, the results demonstrate clear benefits from longer pipeline configurations. On XSum using LLaMA2-13B, PipeSpec with three models \{68M, 7B, 13B\} achieves 2.00$\times$ speedup, significantly outperforming the two-model \{68M, 13B\} configuration at 1.64$\times$. This is also shown for HumanEval using LLaMA3.1-70B, where extending the pipeline from \{8B, 70B\} to \{1B, 8B, 70B\} improves speedup from 2.27$\times$ to 2.54$\times$. These results validate our theoretical analysis showing that pipeline efficiency increases with depth.

\begin{table}[t]
\centering
\caption{Impact of asynchronous pipeline execution and hierarchical model refinement on throughput using LLaMA3.1-70B on HumanEval. Speedup is relative to autoregressive baseline.}
\label{tab:speedup}
\begin{tabular}{lcc}
\toprule
& \multicolumn{2}{c}{Hierarchical Model Pipeline} \\
\cmidrule(l){2-3}
& Single Draft & Multi-Draft \\
\midrule
Synchronous & $1.32\times$ & $1.37\times$ \\
Asynchronous & $2.27\times$ & $2.54\times$ \\
\bottomrule
\end{tabular}
\end{table}

To better understand the contributions of PipeSpec's key architectural innovations, we conducted an ablation study on HumanEval using our LLaMA3.1-70B configuration, shown in Table~\ref{tab:speedup}. Disabling asynchronous pipeline execution (forcing synchronous stage dependencies) reduces speedup from 2.54$\times$ to 1.37$\times$, highlighting the critical importance of breaking traditional stage dependencies. This substantial performance drop aligns with our theoretical analysis in Section 3.3, which predicted that eliminating synchronization overhead would be the primary driver of PipeSpec's advantages over traditional speculative decoding approaches.

Similarly, using only a single draft model instead of our hierarchical pipeline drops performance to 2.27$\times$ under asynchronous execution, demonstrating the value of progressive token refinement through intermediate models. The baseline configuration with both synchronous execution and single draft model (effectively standard speculative decoding) achieves only 1.32$\times$ speedup, validating our architectural decision to pursue both asynchronous execution and hierarchical refinement in the full PipeSpec framework.

Finally, PipeSpec achieves competitive or superior performance compared to more complex algorithmic approaches like LayerSkip~\cite{elhoushi2024layer} and Draft\&Verify~\cite{zhang2023draft}, despite these methods employing sophisticated model-specific optimizations or additional pre-training. For instance, on CNN/DM using LLaMA2-13B, PipeSpec achieves 1.93$\times$ speedup compared to 1.81$\times$ for LayerSkip. Since these methods optimize different aspects of the inference process, they could potentially be combined with PipeSpec's asynchronous pipelining to achieve even greater speedups. (Note that speedup numbers for related works are taken from their original papers, though we use identical verifier model configurations and sizes for fair comparison.)

\begin{table}[htb!]
\centering
\caption{Performance across decoding strategies. Speedup is relative to Autoregressive (AR) baseline. Time is in milliseconds/token. {\color{oursblue}PipeSpec} is our method.}
\label{tab:results}
\begin{threeparttable}
\setlength{\tabcolsep}{3pt}
\begin{tabular}{c|l|l rr}
\toprule
& Method & Models & Time & {\small Speedup} \\
\midrule
\multirow{9}{*}{\rotatebox[origin=c]{90}{CNN/DM}} & AR Basline & {\small LLaMA2-7B} & 20.44 & 1.00$\times$ \\
  & Speculative & 68M,7B & 15.12 & 1.35$\times$ \\
  & LayerSkip & {\small LLaMA2-7B} & \best{--} & \best{1.86}$\times$ \\
  & {\color{oursblue}PipeSpec} & 68M,7B & 14.56 & 1.40$\times$ \\
\cmidrule{2-5}
  & AR Basline & {\small LLaMA2-13B} & 30.02 & 1.00$\times$ \\
  & Speculative & 68M,7B,13B & 20.66 & 1.45$\times$ \\
  & Draft\&Verify & {\small LLaMA2-13B} & -- & 1.56$\times$ \\
  & LayerSkip & {\small LLaMA2-13B} & -- & 1.81$\times$ \\
  & {\color{oursblue}PipeSpec} & 68M,7B,13B & \best{15.54} & \best{1.93}$\times$ \\
\midrule[\heavyrulewidth]
\multirow{11}{*}{\rotatebox[origin=c]{90}{XSum}} & AR Basline & {\small LLaMA2-7B} & 20.55 & 1.00$\times$ \\
  & Speculative & 68M,7B & 15.19 & 1.35$\times$ \\
  & LayerSkip & {\small LLaMA2-7B} & -- & 1.54$\times$ \\
  & {\color{oursblue}PipeSpec} & 68M,7B & \best{12.63} & \best{1.63}$\times$ \\
\cmidrule{2-5}
  & AR Basline & {\small LLaMA2-13B} & 30.26 & 1.00$\times$ \\
  & Speculative & 68M,13B & 20.60 & 1.47$\times$ \\
  & Speculative & 68M,7B,13B & 21.32 & 1.42$\times$ \\
  & {\small Draft\&Verify} & {\small LLaMA2-13B} & -- & 1.43$\times$ \\
  & LayerSkip & {\small LLaMA2-13B} & -- & 1.48$\times$ \\
  & {\color{oursblue}PipeSpec} & 68M,13B & 18.45 & 1.64$\times$ \\
  & {\color{oursblue}PipeSpec} & 68M,7B,13B & \best{15.13} & \best{2.00}$\times$ \\
\midrule[\heavyrulewidth]
\multirow{12}{*}{\rotatebox[origin=c]{90}{HumanEval}} & AR Basline & {\small LLaMA2-13B} & 28.19 & 1.00$\times$ \\
  & Speculative & 68M,7B,13B & 21.99 & 1.28$\times$ \\
  & {\small Draft\&Verify} & {\small CLLaMA2-13B} & -- & 1.46$\times$ \\
  & LayerSkip & {\small LLaMA2-13B} & -- & 1.66$\times$ \\
  & {\color{oursblue}PipeSpec} & 68M,7B,13B & \best{15.50} & \best{1.82}$\times$ \\
\cmidrule{2-5}
  & AR Basline & {\small LLaMA3.1-70B} & 123.69 & 1.00$\times$ \\
  & Speculative & 8B,70B & 93.42 & 1.32$\times$ \\
  & Speculative & 1B,8B,70B & 90.14 & 1.37$\times$ \\
  & {\color{oursblue}PipeSpec} & 8B,70B & 54.52 & 2.27$\times$ \\
  & {\color{oursblue}PipeSpec} & 1B,8B,70B & \best{48.76} & \best{2.54}$\times$ \\
\bottomrule
\end{tabular}
\end{threeparttable}
\end{table}

\subsection{Token Generation Distribution and Timing Analysis}
\label{sec:eval:token-analysis}

\begin{figure}[t]
    \centering
    \includegraphics[width=\linewidth]{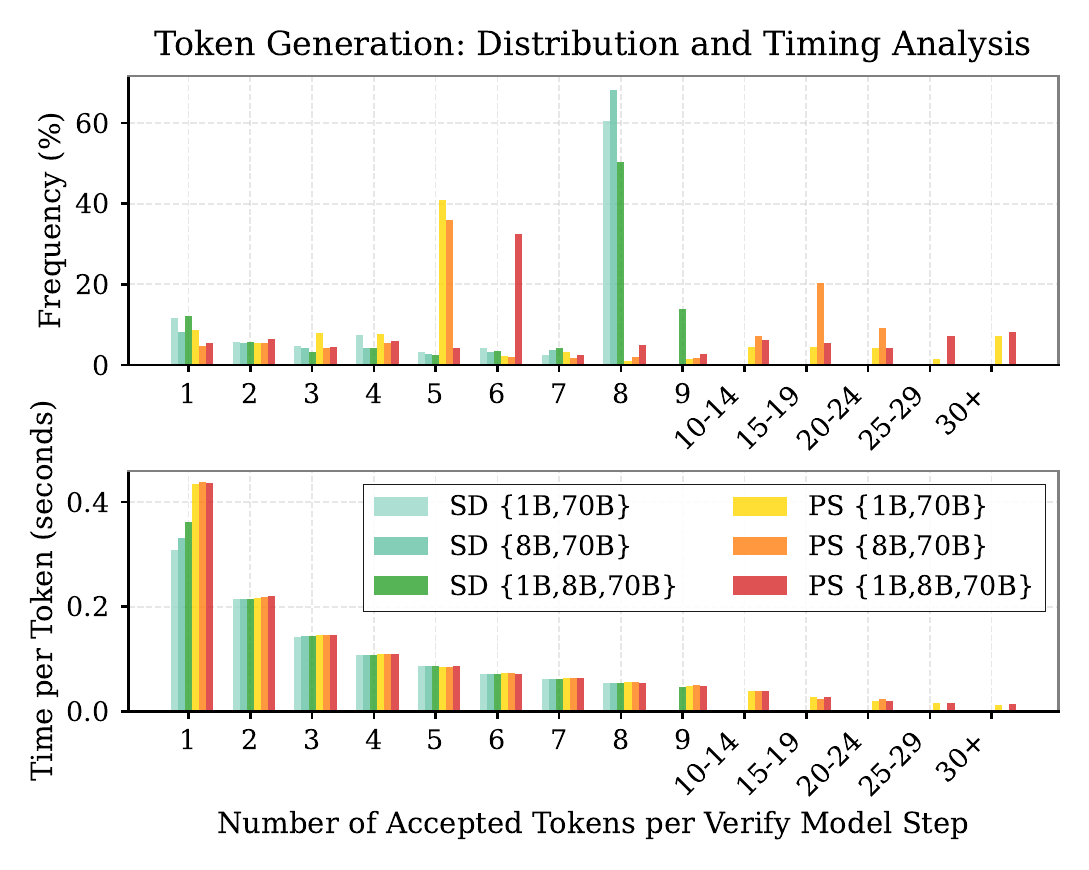}
    \caption{Analysis of token acceptance patterns and timing across decoding strategies on HumanEval. Top: Distribution of accepted tokens per verify step, showing SD's fixed window behavior versus PipeSpec's more flexible patterns. Bottom: Average time per token as a function of batch size, demonstrating PipeSpec's minimal synchronization overhead.}
    \label{fig:token-analysis}
\end{figure}

Figure~\ref{fig:token-analysis} presents a comparative analysis of token acceptance patterns between speculative decoding (SD) and PipeSpec (PS) across different model configurations, aggregated across all samples in the HumanEval dataset. The top portion shows the frequency distribution of accepted tokens per step by the verify model, while the bottom portion shows the average time per token.

SD exhibits a pronounced spike at 8 tokens per verification step across all configurations, resulting from its fixed lookahead window size. This creates a rigid operational pattern where SD must strictly alternate between drafting and verifying batches of 8 tokens, balancing between batch processing efficiency and computational waste.

PipeSpec exhibits a notable long-tail distribution in token acceptance patterns, with successful verifications extending well beyond 20 tokens in both two-model PS \{1B, 70B\} and three-model PS \{1B, 8B, 70B\} configurations. The asynchronous design enables natural acceptance patterns to manifest, with a distinctive spike at 6 tokens in the three-model setup emerging from pipeline stage optimizations. This flexibility, combined with the intermediate model's filtering effect, facilitates larger batch sizes by efficiently discarding lower-quality predictions before they reach the computationally intensive verification stage at 70B. This long-tail distribution indicates that PipeSpec can effectively capitalize on `easy' prediction sequences where models agree, allowing significantly larger sequences to be processed when token predictions align well, while still maintaining fast recovery through the pipeline when predictions diverge.

\subsection{Token Lookahead Analysis}
\label{sec:eval:lookahead}
\begin{figure}[t]
\centering
\includegraphics[width=\linewidth]{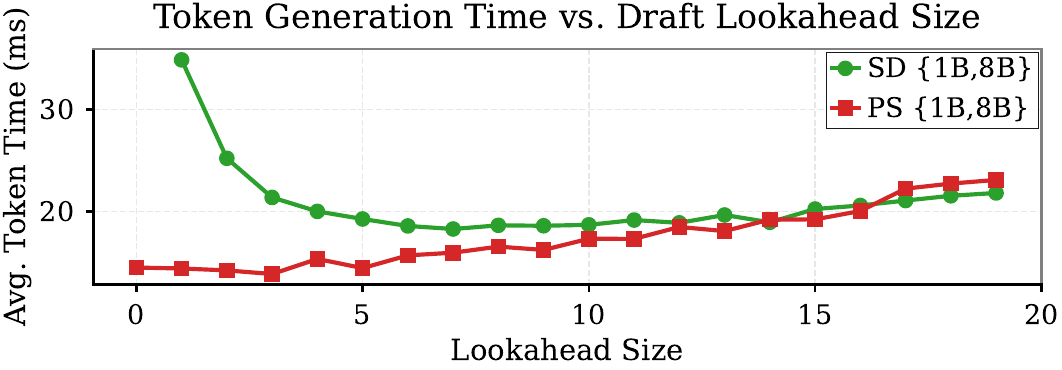}
\caption{Impact of lookahead window size on token generation time. SD shows poor performance at small windows due to synchronization overhead and at large windows due to wasted speculation. PS maintains lower latency at small windows but degrades at larger sizes as verification must wait for draft tokens.}
\label{fig:lookahead}
\end{figure}

As shown in Figure~\ref{fig:lookahead}, the lookahead window size (the number of tokens generated by draft models before verification) significantly shapes the performance characteristics of both approaches. For SD, small windows (1-5 tokens) lead to high latency as the verify model lacks sufficient tokens to batch process effectively, while moderate windows (5-10 tokens) improve performance through better batching before degrading beyond 10 tokens due to increased speculation waste. In contrast, PS maintains lower latency at small window sizes through continuous pipeline processing, though it also experiences degradation with larger windows as verification must wait for more draft tokens to accumulate. These results reveal different optimal operating points. SD performs best with moderate lookahead windows (8-10 tokens), while PS achieves optimal performance with minimal lookahead. We consistently used lookahead sizes of 8 for SD and 0 for PS.

\subsection{Resource Utilization}
\label{sec:eval:util}
\begin{figure}[t]
\centering
\includegraphics[width=\linewidth]{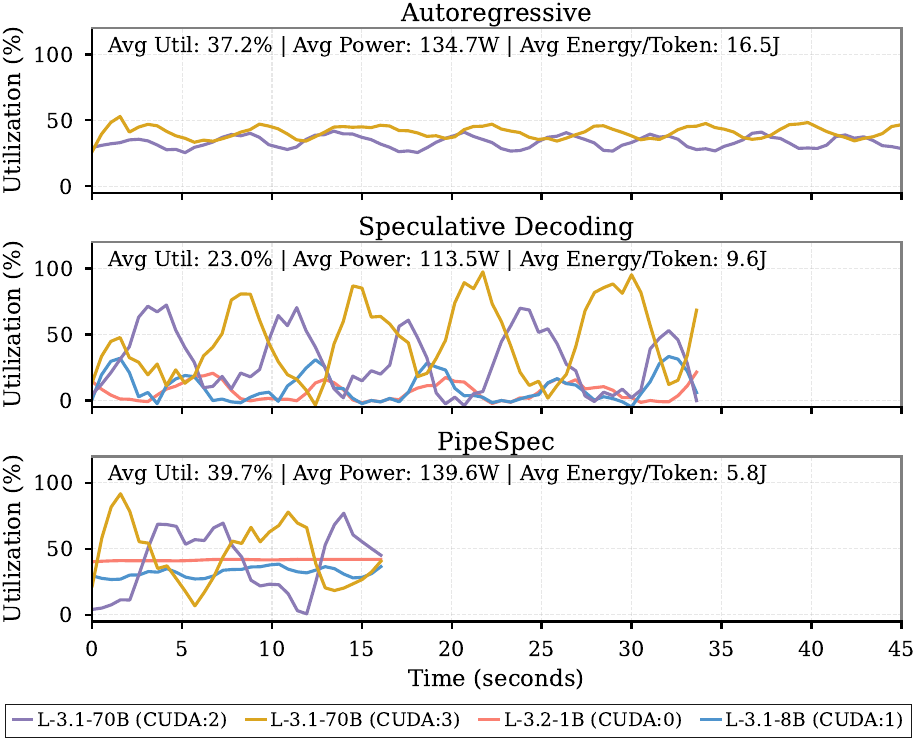}
\caption{GPU utilization over time showing autoregressive (70B model split across 2 GPUs), speculative decoding (\{1B,8B,70B\}), and PipeSpec (\{1B,8B,70B\}). PipeSpec achieves higher average utilization (39.7\%) by eliminating idle periods between draft and verification.}
\label{fig:gpu-usage-analysis}
\end{figure}

Figure~\ref{fig:gpu-usage-analysis} shows GPU utilization patterns across decoding approaches for a HumanEval sample using LLaMA3.1-70B. While autoregressive decoding achieves 37.2\% utilization, traditional speculative decoding exhibits pronounced idle periods where draft models drop to near-zero utilization while awaiting verification, resulting in 23.0\% average utilization. PipeSpec maintains consistently higher GPU activity (39.7\%) through pipelining, eliminating these idle periods. This improved hardware utilization translates to better energy efficiency, with PipeSpec achieving 5.8J/token compared to 16.5J/token for autoregressive decoding.

\section{Conclusion}
\label{sec:conclusion}

We introduced PipeSpec, a novel framework that breaks traditional sequential dependencies in LLM inference through hierarchical pipelined execution. Our theoretical analysis demonstrates that PipeSpec guarantees throughput improvements over autoregressive decoding for any non-zero acceptance rate, while our empirical results show speedups of up to 2.54$\times$ on state-of-the-art models. The key finding that pipeline efficiency increases with model depth—achieving higher speedups with three-model configurations compared to two models—suggests a promising direction for inference architectures that leverage progressive token refinement. By enabling asynchronous execution, PipeSpec provides a scalable approach to accelerating LLM inference.

\newpage
\section*{Limitations}
A key limitation of PipeSpec lies in its static pipeline configuration strategy. The current approach uses fixed model selections and predetermined pipeline depths, which may not be optimal across different tasks or input characteristics. Some generation tasks might benefit from deeper pipelines with more intermediate verification stages, while others might achieve better performance with shallower configurations. The system lacks mechanisms to dynamically adjust its architecture based on task complexity, resource availability, and observed prediction patterns. This rigidity means PipeSpec cannot adapt to changing computational demands or leverage emerging patterns in token generation that might suggest more efficient pipeline arrangements.

From an implementation perspective, the system's performance is heavily dependent on the quality of draft model predictions. While our hierarchical approach helps mitigate poor predictions through progressive refinement, frequent mispredictions can still trigger expensive rollback cascades across multiple pipeline stages. The current design assumes all models can fit within available GPU memory, with larger models split across devices as needed. This may not scale effectively to scenarios with more severe memory constraints or when using very deep pipelines with many intermediate models. Additionally, while PipeSpec reduces overall inference latency, it does so at the cost of increased energy consumption and hardware requirements compared to single-model approaches. The continuous parallel execution across multiple GPUs leads to higher sustained power draw, raising important questions about the trade-offs between speed and efficiency as language models continue to grow in size and complexity.

\bibliography{main}

\end{document}